# Synthetic dual image generation for reduction of labeling efforts in semantic segmentation of micrographs with a customized metric function

Matias Oscar Volman Stern, Dominic Hohs, Andreas Jansche, Timo Bernthaler and Gerhard Schneider

***Abstract*—Training of semantic segmentation models for material analysis requires micrographs as the inputs and their corresponding masks. In this scenario, it is quite unlikely that perfect masks will be drawn, especially at the edges of objects, and sometimes the amount of data that can be obtained is small, since only a few samples are available. These aspects make it very problematic to train a robust model. Therefore, we demonstrate in this work an easy-to-apply workflow for the improvement of semantic segmentation models of micrographs through the generation of synthetic microstructural images in conjunction with masks. The workflow only requires joining a few micrographs with their respective masks to create the input for a Vector Quantised-Variational AutoEncoder (VQ-VAE) model that includes an embedding space, which is trained such that a generative model (PixelCNN) learns the distribution of each input, transformed into discrete codes, and can be used to sample new codes. The latter will eventually be decoded by VQ-VAE to generate images alongside corresponding masks for semantic segmentation. To evaluate the quality of the generated synthetic data, we have trained U-Net models with different amounts of these synthetic data in conjunction with real data. These models were then evaluated using the non-synthetic or real images only. Additionally, we introduce a customized metric derived from the mean Intersection over Union (mIoU) that excludes the classes that are not part of the ground-truth mask when calculating the mIoU of all the classes. The proposed metric prevents a few falsely predicted pixels from greatly reducing the value of the mIoU. As a result of the developed workflow, we have achieved a reduction in sample preparation and acquisition times, as well as the efforts, needed for image processing and labeling tasks, are less when it comes to training semantic segmentation model. The idea behind this work is that the approach could be generalized to various types of image data such that it serves as a user-friendly solution for training models with a smaller number of real images.**

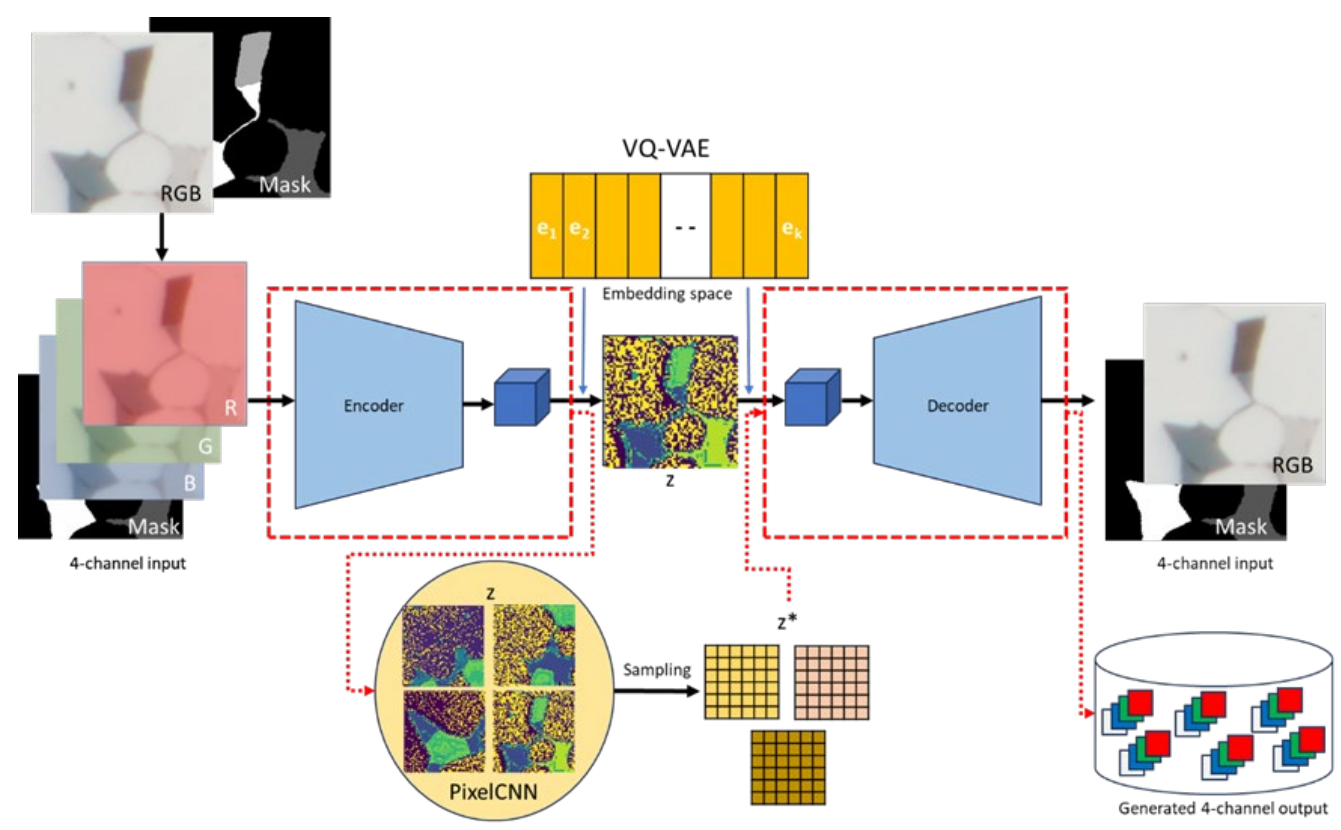

Fig. 1. Graphical abstract. Combination of VQ-VAE and PixelCNN models for synthetic dual image generation. At the top, the VQ-VAE layout with 4-channel input and output images. The red lines represent the second part of the workflow, where the trained VQ-VAE encoder produces discrete latent representations (z-matrices) that are used to train a PixelCNN model. The latter model is used to sample z* that are fed into the trained VQ-VAE decoder to generate dual images (RGB micrograph + grayscale mask).

## I. Introduction

CREATING semantic segmentation models for quantitative microstructural analysis is often an arduous task. One of the most time-consuming and difficult tasks to perform is the labeling of images, which, according to [1], requires in minimum approximately 25% of the total time of any machine learning (ML) project. Besides, the creation of perfect masks, pixel by pixel, is quite unlikely to be achieved, especially at the edges of the objects (phases in material science or semantic classes in computer science). An important aspect to highlight as well is the amount of data that can be obtained, many times, only one or a few samples are available, which makes it very problematic to train a robust deep learning (DL) model.

Until today, when there is the necessity to populate datasets, data augmentation is commonly applied or other methods that produce small variations in comparison to the original images. An alternative used option, not that common, is image-to-image translation, for example using the Pix2Pix [2] architecture, but these models need realistic masks to output microstructural

Preprint submitted to arXiv on 01.08.2024. This work was funded by the Federal Ministry of Education and Research Germany (BMBF) in the scope of the WirLebenSOFC project (grant no. 03SF0622C).

The authors are with the Materials Research Institute Aalen, Aalen University of Applied Sciences, Beethovenstraße 1, 73430 Aalen, Germany. (e-mail: matias_oscar.volmanstern@hs-aalen.de)

images and again obtaining good masks is a long and complex process.

In this work, we present a workflow to synthetically generate images of microstructure together with its respective mask. This workflow differs from data augmentation and image-to-image translation approaches, as it does not focus on generating images with global changes over the original image, but on generating completely new images.

The aim is to investigate whether the use of these images can benefit the training of semantic segmentation models and thus reduce labeling and preparation time significantly. For this purpose, we used only one large 3-channel brightfield image of a $Fe_{14}Nd_2B$ magnet microstructure with the corresponding label to train a VQ-VAE [3] model combined with a PixelCNN [4] model, in order to generate synthetic 4-channel outputs, composed of image-mask pairs.

Once the images are generated, several datasets are prepared, with different percentages of synthesized images, to train U-Net [5] models and, to evaluate those models with real test data only. Noting that common metrics such as accuracy or mIoU were not effective in evaluating these models, we developed a derivative of the mIoU metric. The results indicate that models with synthetic images are likely to have performance improvements when looking at this customized metric that we call “Missing Class IoU”.

## II. Related work

Although there have been more than a few approaches [6] [7] for the generation of synthetic microstructural images, the generation of those in conjunction with the mask is not a common task in this context. One area where DL-models for semantic segmentation are being developed is medicine. In this case, more attempts have been made to create models that generate medical images along with their corresponding masks.

For example, T. Neff et al. [8] used a modified Generative Adversarial Network (GAN) architecture to generate the image-mask pair through random noise. They adapted the Deep Convolutional GAN to work with 2-channel input and trained on a small dataset containing 30 images. While their results using a combination of real and synthetic images did not show any improvement in binary segmentation, these were comparable to those using only real data.

Another procedure developed in this latter field is SinGAN-Seg [9]. The authors mention other methods for generating synthetic data but those contain a lot of manual steps, such as modifications of the real images, mask generation, image processing, etc. So, they develop a method that allows to generate multiple 4-channel synthetic images from a real one with the same number of channels without any manual modification.

The process consists of the following: in the first step, a training of a model per image to generate the synthetic images including the binary mask and then, a style transfer of features by image, such as texture, from the real to synthetic ones. In their experiment, improved performance of the segmentation models was demonstrated when trained on small datasets.

One of the drawbacks of using this procedure is to train a model for each image in the dataset. However, with their methodology, the synthetically generated medical images can be shared with other institutions or laboratories, whereas for real images this is usually restricted, so they achieved one of their goals in developing this type of solution.

P. Andreini et. al. [10] proposed different methods to generate synthetic images with their corresponding masks in medical datasets. The authors applied three approaches, based on the Progressively Growing Generative Adversarial Network (PGGAN) [11] and some of them with a combination of Pix2PixHD [12], and demonstrated that, in this way, they could match or improve some state-of-the-art semantic segmentation methods.

Our workflow is based on VQ-VAE, which was developed to gather important features of the data and represent those in a discrete latent space structure (z-matrices). This method is completely unsupervised and has demonstrated that its latent space can be used to generate meaningful data such as video, images and text.

Next, we use PixelCNN, similar to how these authors used it, as an autoregressive model that is trained on z-matrices for the generation of new synthetic z-matrices. One of the drawbacks of this architecture is the "blind spot" problem, but there are more advanced architectures that overcome this problem, e.g. Gated PixelCNN or other versions of these models that also improve generation performance such as PixelCNN++ [13], but for our experiments, PixelCNN was sufficient.

The model chosen to evaluate the generated synthetic images is U-Net, which was developed for binary segmentation of biomedical images, although it is well suited to many semantic segmentation tasks. A metric commonly used to evaluate binary segmentation models is the IoU [14], or in multi-class cases, the mIoU that calculates the IoU per class and then averages it.

## III. Workflow preparation

### A. *Material / data*

The material studied in this paper is a brightfield extended depth-of-focus (EDF) image of a $Fe_{14}Nd_2B$ magnet [15] (Fig. 2). This magnet is made up of five phases, and the mask was inferred using a semantic segmentation model trained by a magnet expert, whose dataset consisted of other types of magnet images.

Note that some over segmentation and mispredictions occurred, especially at the edges, due to the low quality of the EDF image. Therefore, the mask is not completely accurate, as it happens in most cases when working with microstructures in a lab setting or routine workflow.

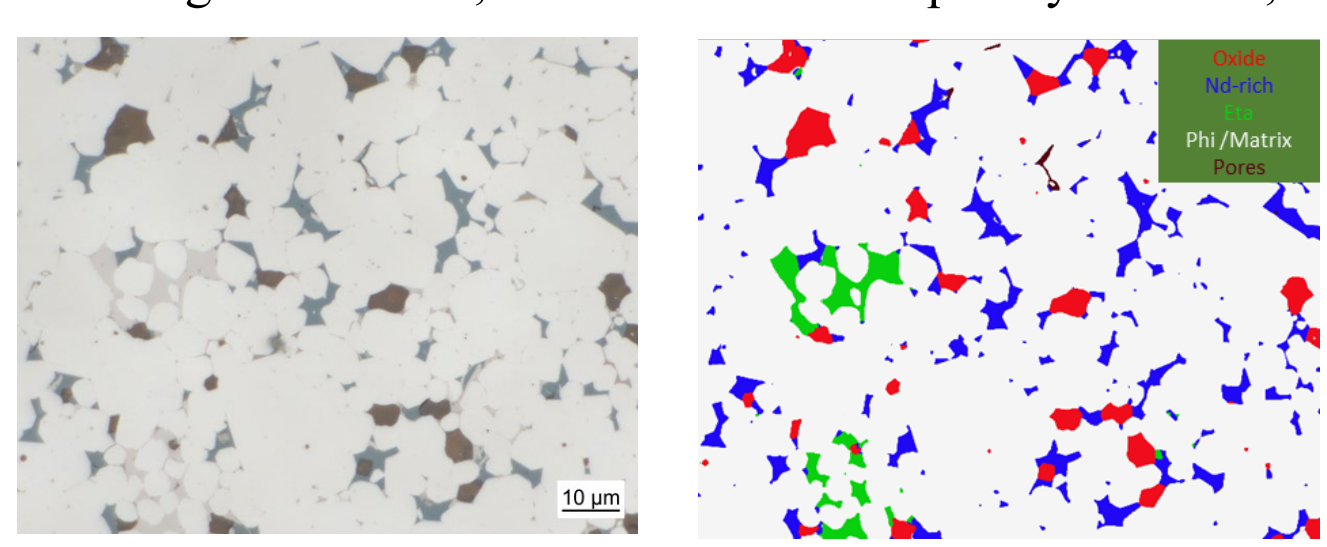

Fig. 2. Left: 2752x2208px EDF image of a $Fe_{14}Nd_2B$-magnet made with optical microscopy and brightfield illumination. Right: Its corresponding mask obtained by a DL-model trained by a magnet expert.

For the first part of the experiment, the synthetic generation, we prepared the data as follows. The large EDF image was cropped to create 80 patches of 256x256px. Of the 80 patches, 16 were originally segmented with pores, but upon image review, only 2 of them contained actually small pores, so we ignored the latter to simplify the workflow. Of the remaining 14, 10 were discarded because they contained doubtful objects and were difficult to label and 4 were relabeled by removing the incorrect pore regions and enhancing those of the other classes, to become part of the dataset.

Of the original 64 pores-free patches, we reviewed each of their masks in an attempt to improve their quality. However, for several patches, a relabeling was difficult to make, as they contained some regions where magnet experts could not distinguish between phases, mainly due to the image quality.

Therefore, we only relabeled 23 patches that were easy to do, since only a few pixels were corrected due to over-segmentation or, in other cases, mislabeled pixels of clear objects were modified to belong to the correct class. The goal of this relabeling was to train DL-models with better quality inputs but as realistic as possible, since those that were not relabeled simulate low-quality masks that are typically found in most real-life datasets.

When creating the square patches, some regions with clear objects at the edge of the original image (rectangular shape) were not used, so 1 more patch that did not contain pixels overlapping with other patches was extracted, labeled and used in the dataset. Therefore, the complete dataset consisted on a total of 69 patches, of which 29 were manually relabeled and became part of the SHQ dataset.

TABLE I shows the four initial datasets that were used in different ways as the basis for other ones, depending on the method required (see section IV.). Where LQ (Low Quality) refers to the set of labels extracted from the segmented image, HQ (High Quality), only relabeled/corrected ones and MQ (Medium Quality), both types of masks. Note that the test set is always the same and is only composed of corrected masks

### B. *Image preprocessing*

In case of VQ-VAE, we combined the RGB micrograph with the grayscale mask to create a 4-channel image and for semantic segmentation, the micrograph was separated from the mask (Fig. 3). On both cases, the masks were converted from RGB to grayscale.

TABLE I
INITIAL DATASETS

| Dataset name | Train data | Test data | Information of train data |
|---|---|---|---|
| SHQ | 21 | 7 | Manually relabeled |
| SLQ | 21 | 7 (SHQ) | Segmented mask |
| BLQ | 62 | 7 (SHQ) | Segmented mask |
| BMQ | 62 (21 SHQ) | 7 (SHQ) | Replacement - SLQ → SHQ |

SHQ: Small High Quality, SLQ: Small Low Quality, BLQ: Big Low Quality, BMQ: Big Medium Quality

### C. *Workflow for experiment execution*

In order to effectively develop the study, it was structured as shown is Fig. 4.

With the BMQ training dataset, we trained the VQ-VAE and PixelCNN models to generate synthetic images of microstructures together with masks. The VQ-VAE model was trained to achieve the lowest reconstruction loss and to produce the z-matrices/codes in the embedding space while the PixelCNN model, to learn the distribution of those matrices and generate new ones. Then, we decoded the generated matrices with the trained VQ-VAE decoder and split the resulting images into microstructure and mask.

To pre-test the quality of the synthetic images, we took the microstructural images to be inferred using the domain expert's trained segmentation model. Once the model issued the masks, a visual comparison with the synthetic ones was carried out. The idea of this evaluation was to see if the two masks were similar; otherwise, we could assume that the synthetic images were of poor quality and the model was to be retrained.

After the successful evaluation, starting from the initial datasets (TABLE I) as a reference, we arranged new diverse datasets including different amounts of synthetic data to analyze how these data influence the predictions of the semantic segmentation models. For this purpose, 46 U-Net models were trained for a quantitative and qualitative analysis of the synthetic data.

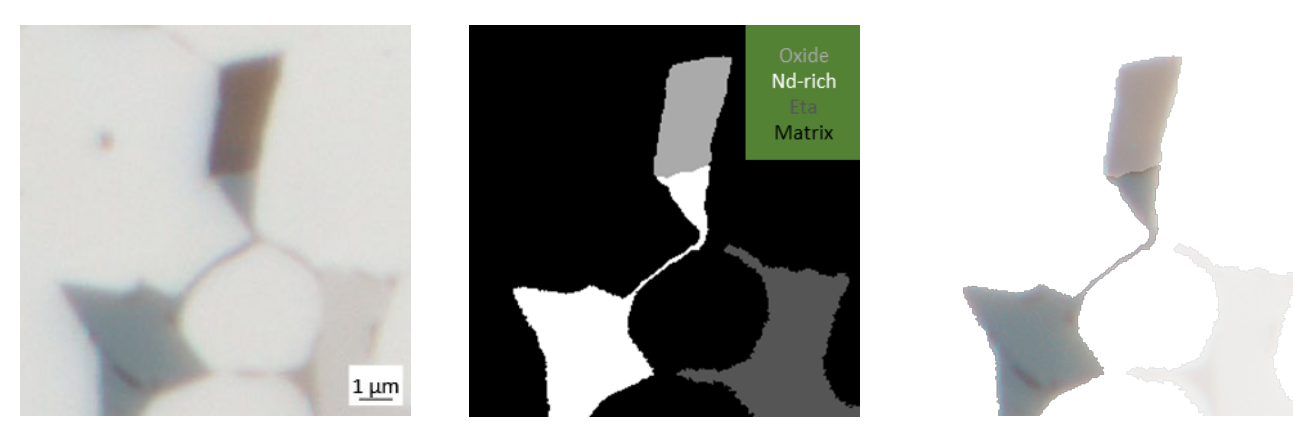

Fig. 3. Left: 256x256px RGB patch from the $Fe_{14}Nd_2B$-magnet image. Center: Mask in grayscale. Color assignments: Light gray – Oxide, White – Nd-rich, Eta – Gray, Black – Matrix. Right: 4-channel image (combination of the RGB microstructure and mask).

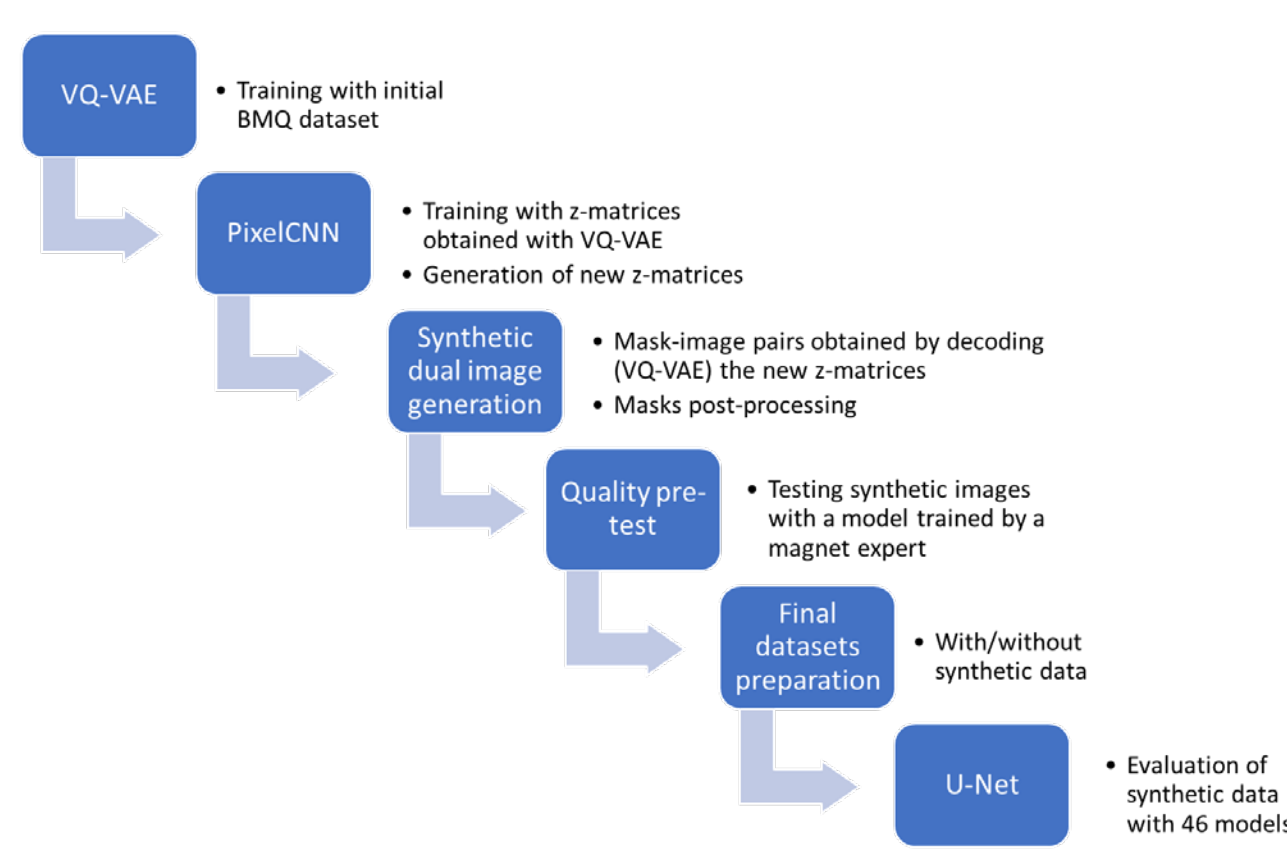

Fig. 4. Flow chart of the main experiment with the $Fe_{14}Nd_2B$-magnet dataset. For the other experiments, only the first three steps of this diagram were considered with minor variations, e.g. initial dataset, post-processing, etc.

## IV. Applied Machine Learning Models

### A. *VQ-VAE: Modifications and preliminary results for final datasets preparation*

The implementation of this model was taken from [16] and adapted to the goal of this paper. The most prominent change in this work was the combination of the micrographs with the masks to be the input of the encoder, as in turn, the decoder was adjusted to provide an output with the same dimensions (Fig. 6).

Several tests were conducted to find the model hyperparameters that fit this approach and the main modifications with respect to the authors' example were the following ones:

For training the model, we selected the BMQ dataset since we could not achieve favorable results with the SHQ. Of the 62 patches, 56 were part of the training and 6 for validation, then as the number of available images was much smaller than the original example from the authors, the batch size was reduced to 8.

Apart from that, since the selected images had 3 classes/phases and the background (matrix), the embedding vectors were significantly reduced, so that the PixelCNN model could learn the matrix distribution of the embedding space by a combination of a reduced number of vectors while the VQ-VAE decoder could still reconstruct good quality outputs (Fig. 7).

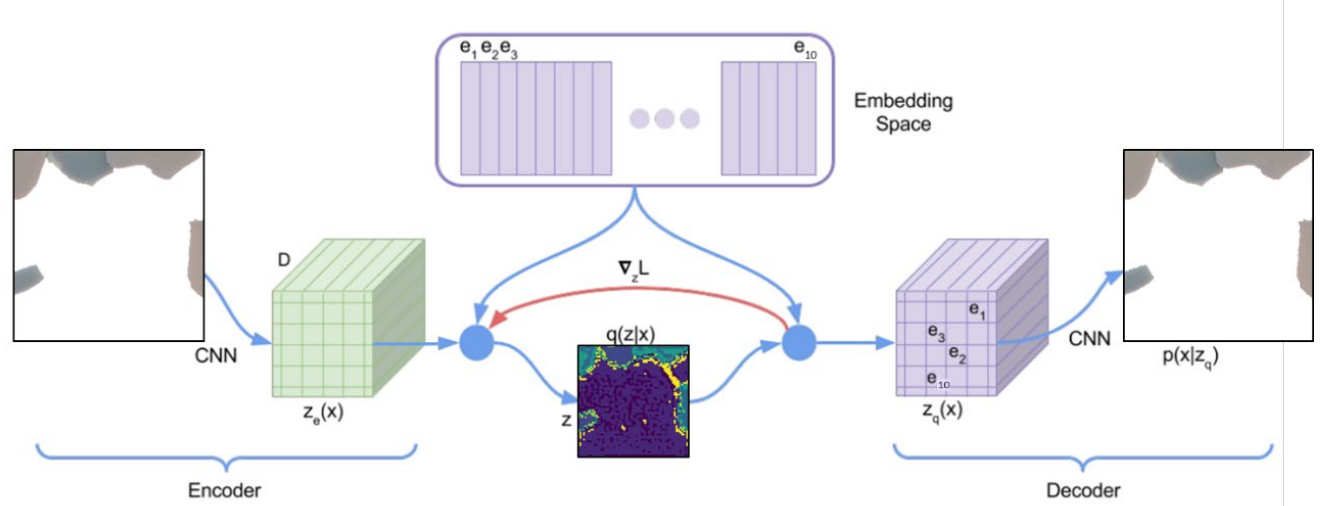

Fig. 6. Training scheme of VQ-VAE extracted from [3] and modified with the paper approach. The differences that can be seen in the figure are: K=10, the 4-channel images as input and output and the matrix z containing values from 1 to 10.

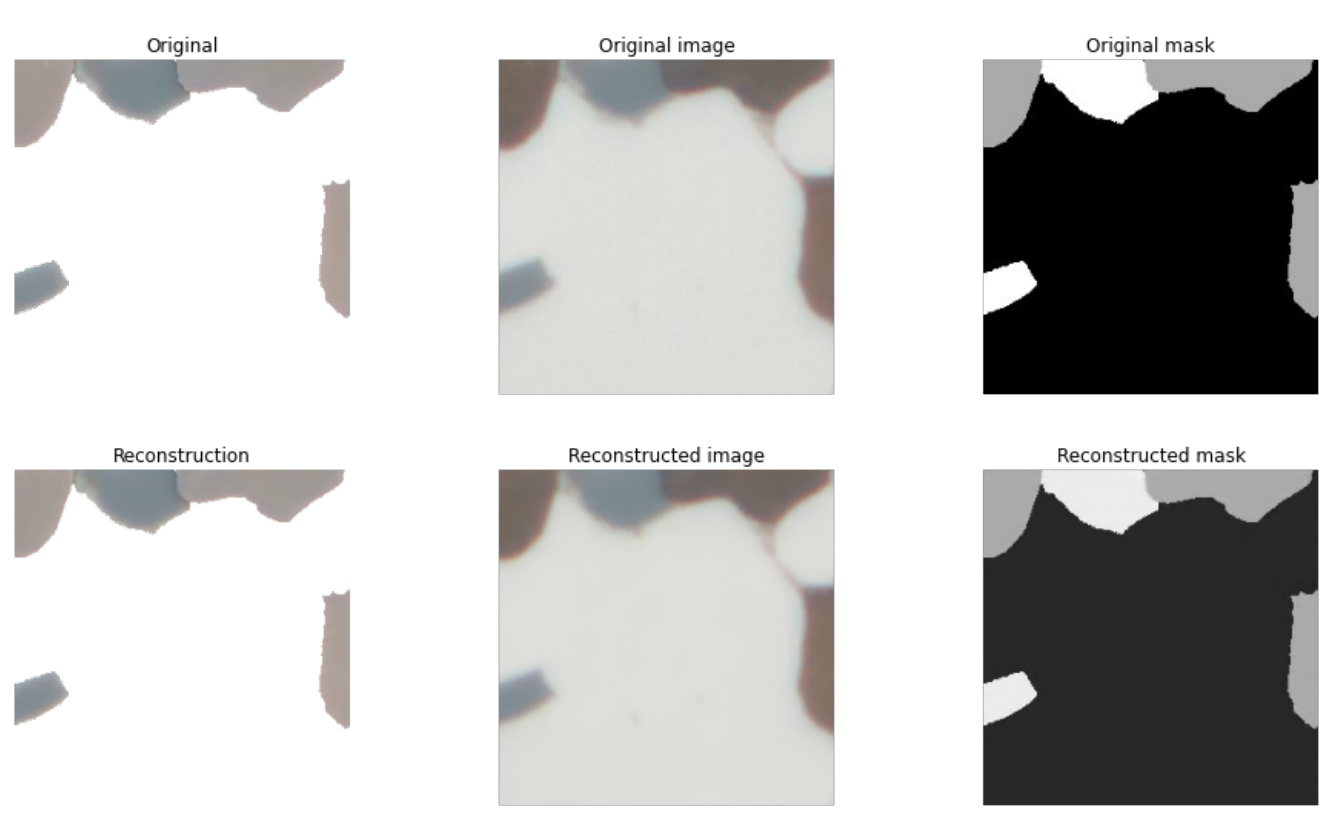

Fig. 7. Top: 256x256px original image taken from the dataset. Bottom: 256x256px reconstruction using the trained VQ-VAE model.

TABLE II
Modification of the Model Hyperparameters in the VQ-VAE Model

| Hyperparameters | DeepMind [16] | Ours |
|---|---|---|
| Dataset and input shape (Height, Width, Channels) | (32,32,3) | (256,256,4) |
| Number of images | 60000 (Cifar10) | 62 (BMQ) |
| Batch size | 32 | 8 |
| D (embedding dimension) | 64 | 64 |
| K (embedding vectors) | 512 | 10 |
| Number of training updates | 10000 | 50000 |

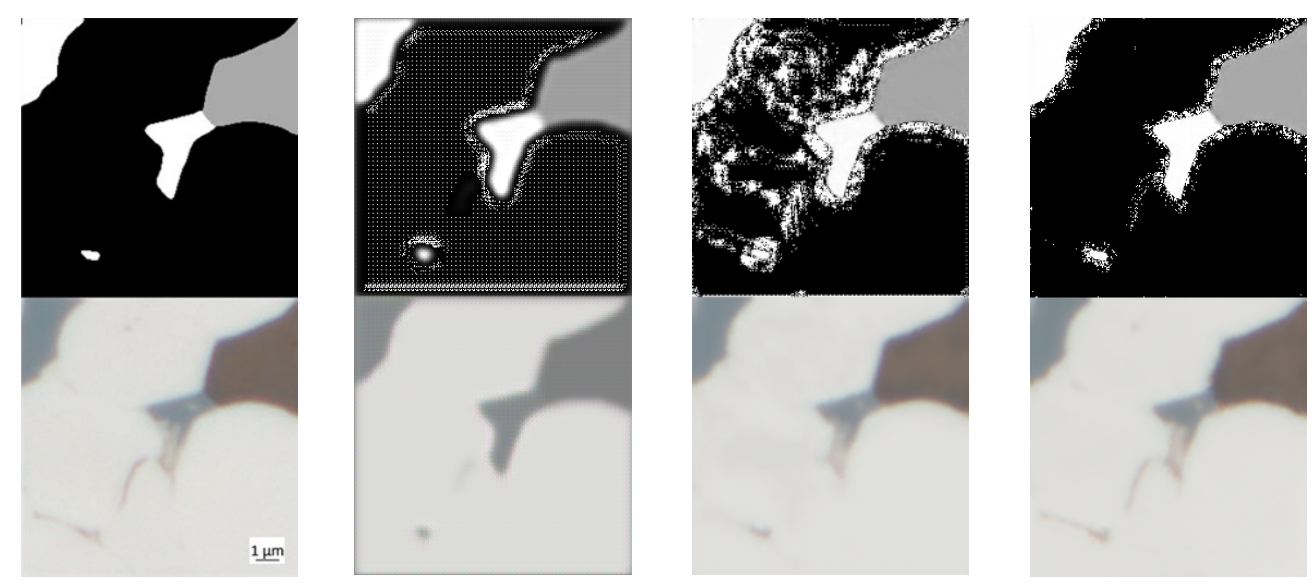

| Validation | Reconstruction | Reconstruction | Reconstruction |
|---|---|---|---|
| **Step:** | 100 | 120000 | 46700 |
| **Recon_error:** | 0.239 | 0.002 | 0.001 |

Fig. 5. VQ-VAE training monitoring. From left to right: 1. Image from the validation dataset divided into microstructure and mask (The dataset does not contain scalebar). 2-4. Reconstructed images, at different steps of the training, with the corresponding reconstruction error.

For an improvement in the reconstruction error and the quality of the decoded output, we increased the training steps. In addition, the checkpoint was added to the source code, in order to save the model with the lowest reconstruction error.

Step-by-step training of the model was followed by applying inference on an image from the validation dataset at every checkpoint. In Fig. 5, it is shown how at the first checkpoint (Step 100), the reconstruction was composed of horizontal and vertical lines, and almost no differentiation between the phases was visible.

Then at 12000 steps, the model reconstructed the mask very poorly and some regions of the microstructure were not reconstructed. Finally, at the last checkpoint, the model reached the best reconstruction, although there were some errors in the output mask, probably due to artifacts or quality of the images, these did not preclude the aim of this study.

### B. PixelCNN: Modifications and preliminary results for final datasets preparation

Once the VQ-VAE model was trained, we obtained the 62 z-matrices (Fig. 8) composed with the codebook indices from the images of the BMQ dataset to train the PixelCNN model.

At the end of the PixelCNN training, a sampler model was built to generate new z-matrices from empty matrices. This model took empty matrices of 64x64px as input and predicted each element based on pixels that are above and on the left of it. Then, by using the decoder of the VQ-VAE, synthetic 4-channel images were generated. To better visualize them, in Fig. 9 they were split into microstructural image and mask.

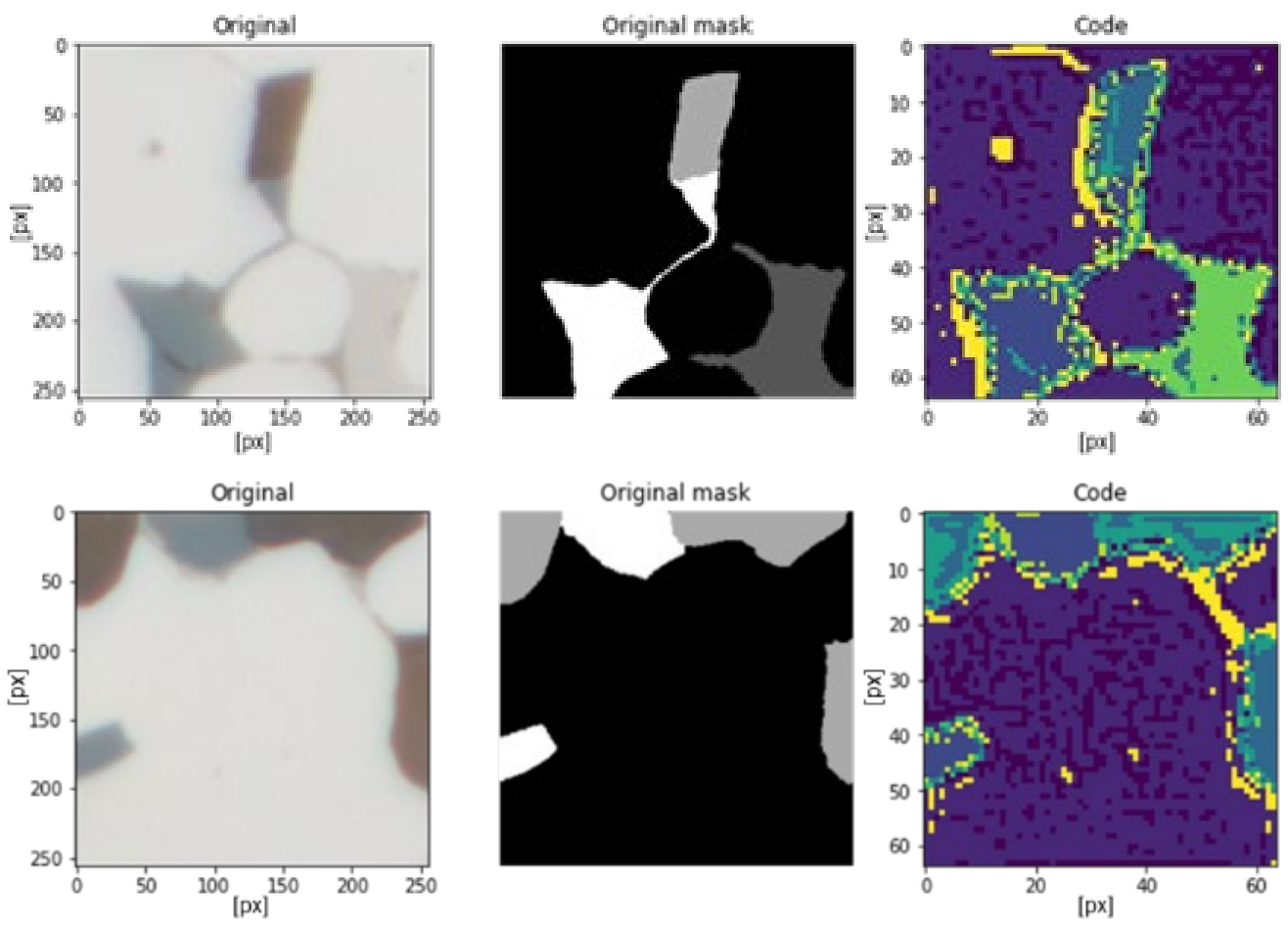

Fig. 8. $Fe_{14}Nd_2B$-magnet - Representation of images in the trained embedding space. Left: Original microstructural images taken from the dataset. Center: Original masks taken from the dataset. Right: z-matrices composed with the 10 codebook indices.

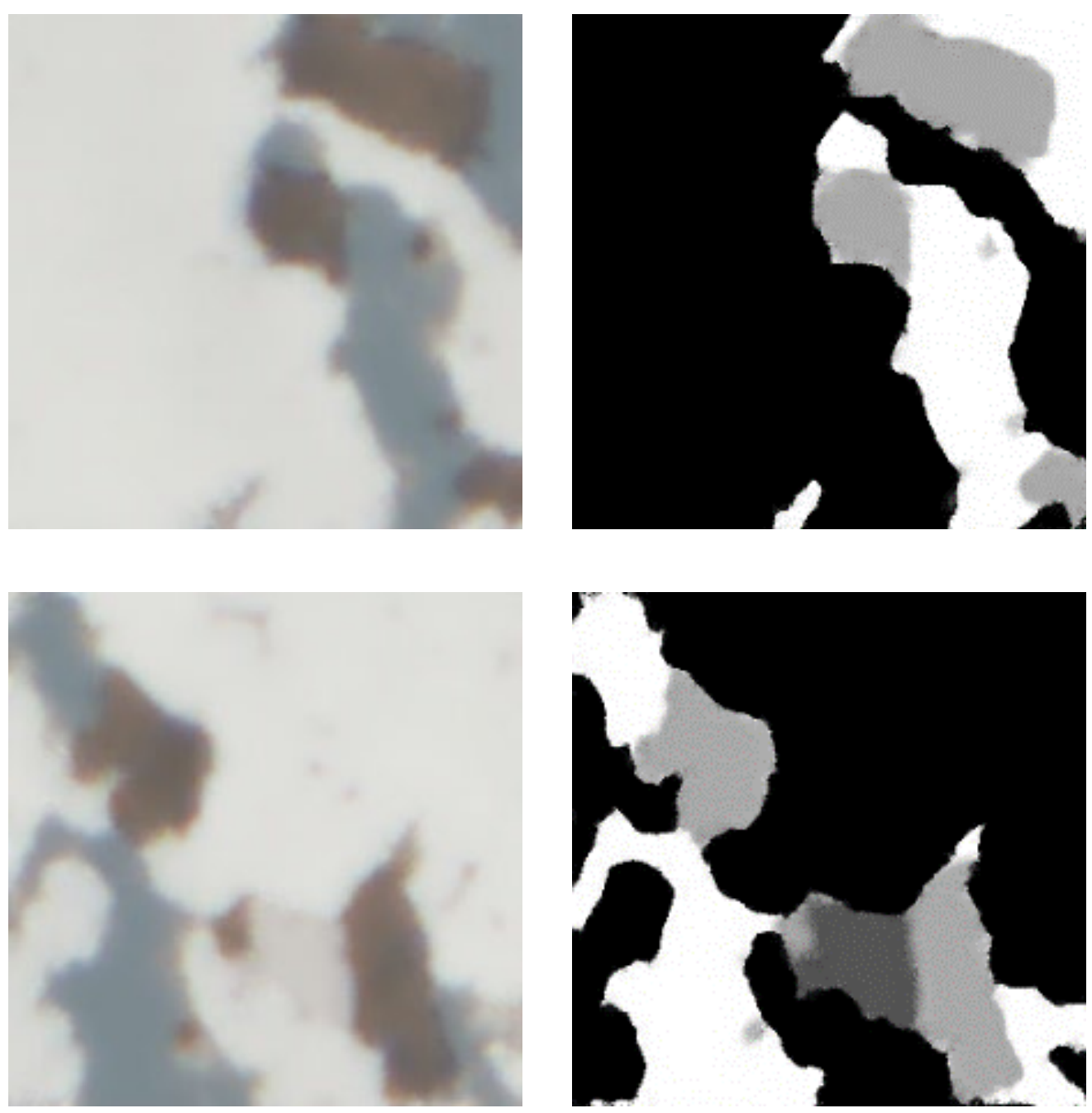

Fig. 9. $Fe_{14}Nd_2B$-magnet - Synthetic image generation. Left: Synthetic microstructural image. Right: Synthetic mask without post-processing.

At first glance, it can be seen that the masks were created correctly, but when observing the boundaries, different pixel values and not just one per class can be identified. This was the result of using poor quality images, so without processing, these masks could not be used directly in semantic segmentation models. To create useful masks for the U-Net, the generated ones were processed. First, we applied the k-means [17] algorithm to group the pixels into the 3 classes and the background; then, we filtered and removed small objects (<200px$^2$), and those regions were filled with the neighboring class (Fig. 10).

A quality pre-test was performed. Fig. 11 shows the raw synthetically generated masks compared to the masks inferred by the subject matter expert model and it can be seen that both are similar, meaning that the model correctly predicted the synthetic objects.

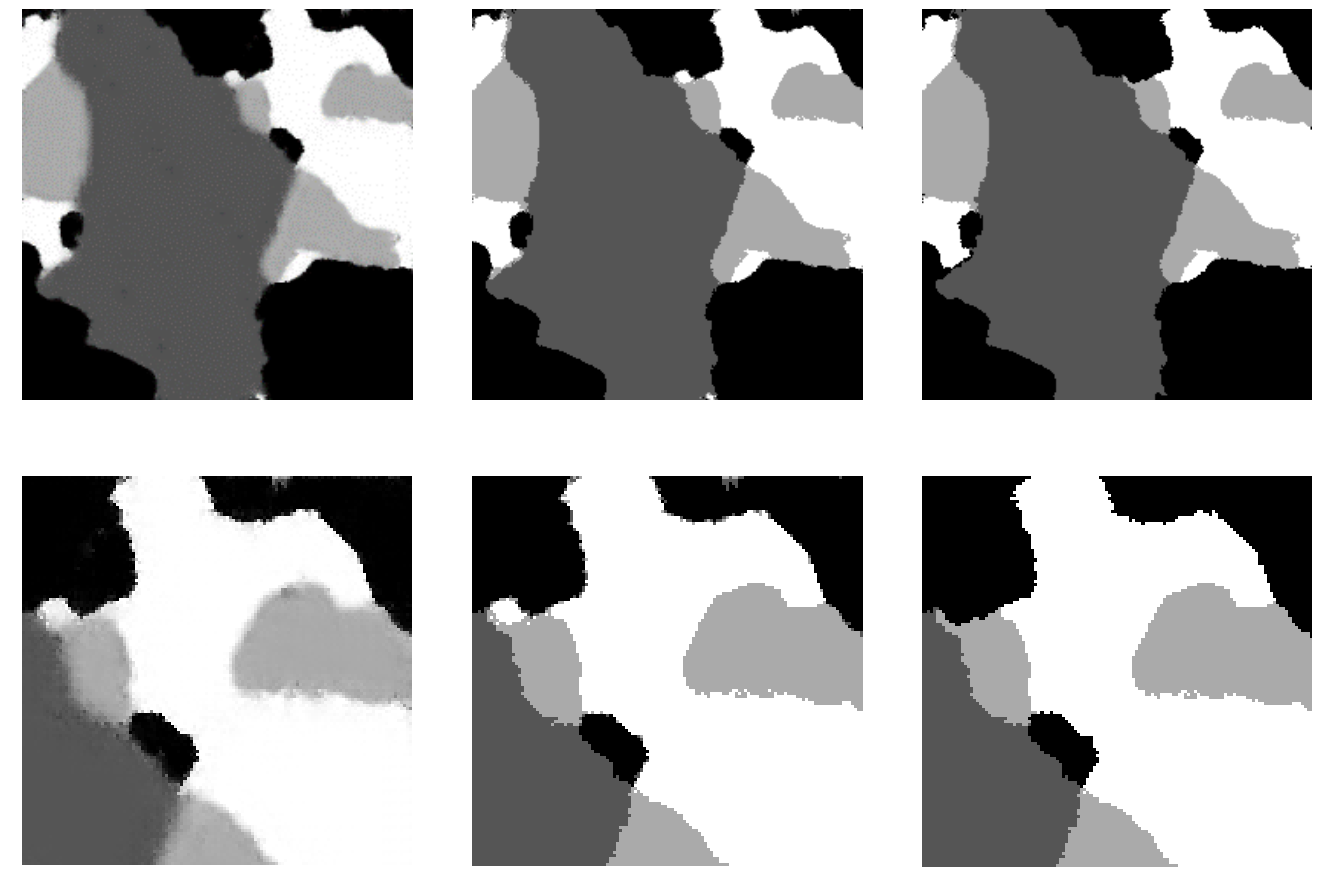

Fig. 10. Step-by-step mask pre-processing for U-Net. Left: Synthetic generated mask. Center: Result from the k-means (4 clusters). Right: Result of filtering small objects and filling them with the correct class.

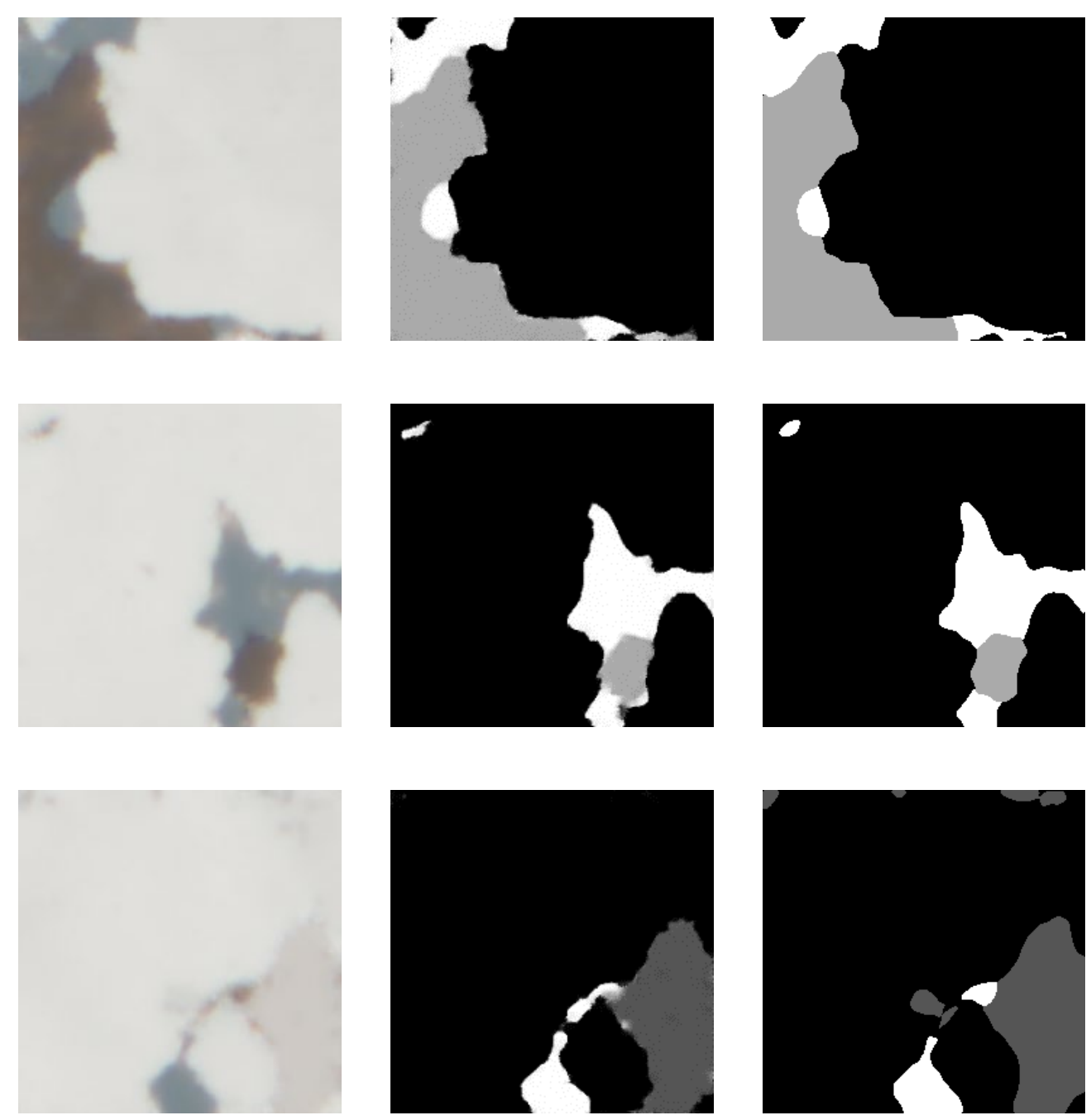

Fig. 11. $Fe_{14}Nd_2B$-magnet - Quality pre-test of synthetic images. Left: Synthetic microstructural image. Center: Synthetic mask without post-processing. Right: Mask inferred by the model of the expert.

*1) Dataset preparation with addition of synthetic data for training semantic segmentation models*

Since, after the generation of the synthetic data, we received more large images of the magnet type studied in this work, 20 well-labeled patches of different images were added to the test dataset. Ideally, the larger the test dataset, the better its evaluation.

It should be noted that the large datasets (BLQ and BMQ)

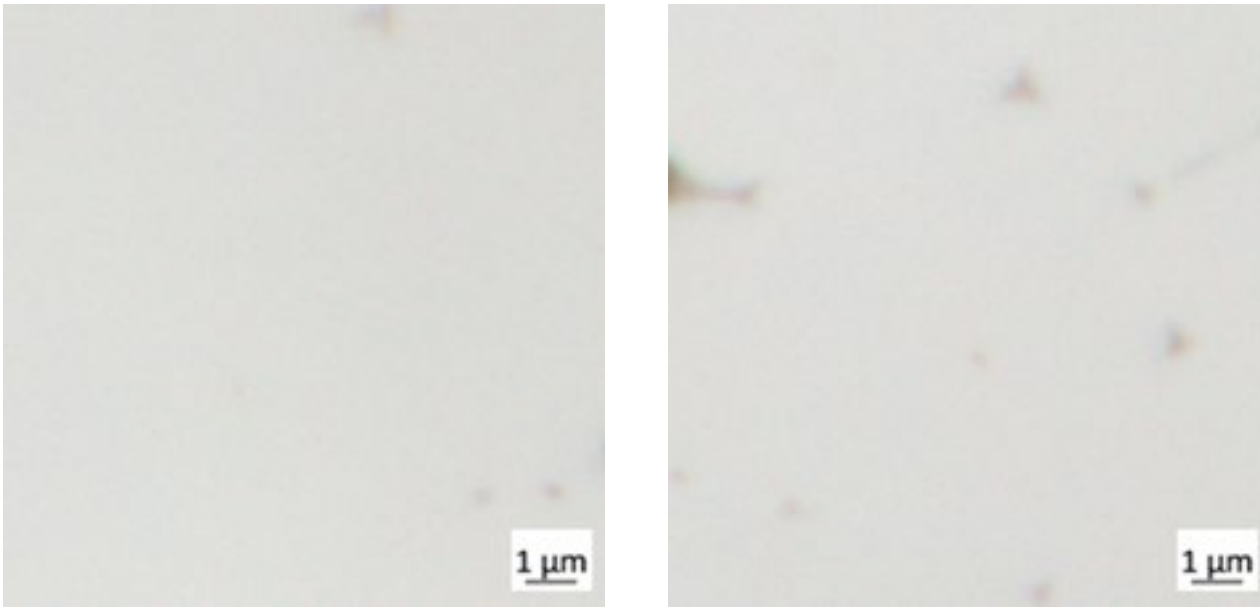

Fig. 12. $Fe_{14}Nd_2B$-magnet - Images from the initial dataset that were discarded for training the semantic segmentation models since no relevant information was present.

TABLE III
PREPARED DATASETS FOR TRAINING THE SEMANTIC SEGMENTATION MODELS

| Dataset name | Train / Val | Addition of synthetic data to the train dataset [%] | Test |
|---|---|---|---|
| SHQ | 16/5 | 50, 75, 100, 150, 200, 250, 300 | 7 + 20 |
| SLQ | 16/5 | 50, 75, 100, 150, 200, 250, 300 | 27 (SHQ) |
| BLQ | 48/12 | 50, 75, 100, 150, 200, 250, 300 | 27 (SHQ) |
| BMQ | 48/12 (21 SHQ) | 50, 75, 100, 150, 200, 250, 300 | 27 (SHQ) |

TABLE IV
PREPARED DATASETS FOR TRAINING THE SEMANTIC SEGMENTATION MODELS

| Dataset name | Train / Val | % Training synthetic data | Test |
|---|---|---|---|
| SyntLQ | 16/5 | 50, 75, 100, 150, 200, 250, 300 | 27 (SHQ) |
| SyntMQ | 48/12 (21 SHQ) | 50, 75, 100, 150, 200, 250, 300 | 27 (SHQ) |

Training data is synthetic. Validation LQ and MQ are real images without synthesized images contained two images less than the initial datasets because they were discarded as no relevant objects were present (Fig. 12).

TABLE III illustrates the initial size of the training and validation datasets in the second column. The training datasets were increased with each value in the third column, while the validation one remained unchanged. The base datasets were based on those described in TABLE II. The columns "Train / Val" and "Test" contain the number of real images, while the third column lists the percentages of synthesized images added to the training dataset with respect to the length of the base training datasets.

For example, from the SHQ training dataset, an SHQ [50%] dataset was created by adding 8 synthetic images (50% of 16). In total, we formed 28 datasets with synthetic data from the initial 4, without synthetic images.

For a more extensive study, only synthesized data were used to create 14 more datasets by considering the size of the large datasets (TABLE IV). The validation datasets were composed only from original images with respect to BLQ and BMQ. As an example, the SyntLQ [50%] and SyntMQ [50%] consisted of the following: Training – 24 synthetic images, Validation – 12 real images (from BLQ and BMQ, respectively) and Test – 27 real images from SHQ.

In other words, the validation and test datasets remained unchanged and contained only real data, while the training datasets were purely synthetic. The values in the second column refer to SyntLQ [100%] and SyntMQ [100%].

TABLE V
SELECTED MODEL HYPERPARAMETERS FOR TRAINING THE 46 U-NET MODELS

| Hyperparameters | Value / Name | Remark |
|---|---|---|
| Backbone | EfficientNetB3 | - |
| Number of classes | 4 | The 3 phases and the matrix (Red, Green, Blue and Black). |
| Input shape | 256x256x3 | - |
| Initial weights | ImageNet | - |
| Epochs | 200 | With "EarlyStopping" to stop training when the IoU score did not improve after 15 epochs. |
| Steps per epoch | Not constant | Length of the train dataset divided by the batch size. |
| Validation steps | Not constant | Length of the validation dataset. |
| Batch size | 2, 1 | Train, validation. |
| Optimizer | Adam with LR= 0.001 | With "ReduceLROnPlateau" callback in order to reduce the learning rate when the IoU did not improve after 5 epochs. |
| Compound loss function | Dice-Focal (Categorical) | - |

### *C. U-Net for quantitative and qualitative analysis of generated images*

We set up a Jupyter Notebook to train the U-Net architecture on each of the 46 datasets in the most equitable manner. The implementation from the GitHub repository [18] was adapted to this work. We kept all model hyperparameters constant (TABLE V) and the only variation was the number of images that made up each dataset.

The purpose of training 46 models with the different datasets was to evaluate the metrics applied on the SHQ test dataset composed of 27 real images and to draw conclusions about the use of synthetic data in semantic segmentation models.

#### *1) Missing Class IoU – A customized metric for model evaluation*

In addition to the common metrics, we developed a personalized metric, "Missing Class IoU", to evaluate the semantic segmentation models. The principle of this new metric is the same as the mIoU but it does not consider classes that are not part of the ground-truth mask when calculating the average IoU of all the classes ($C_n$).

$$Missing\ Class\ IoU = \frac{\sum_i^n IoU_{C_n}}{n} \qquad (1)$$

Thus, in the case of the example given in Fig. 13, the new metric is calculated as follows:

$$\frac{IoU_{Yellow} + IoU_{Blue} + IoU_{Background}}{3} = \frac{1 + \frac{1}{2} + \frac{32}{34}}{3}$$

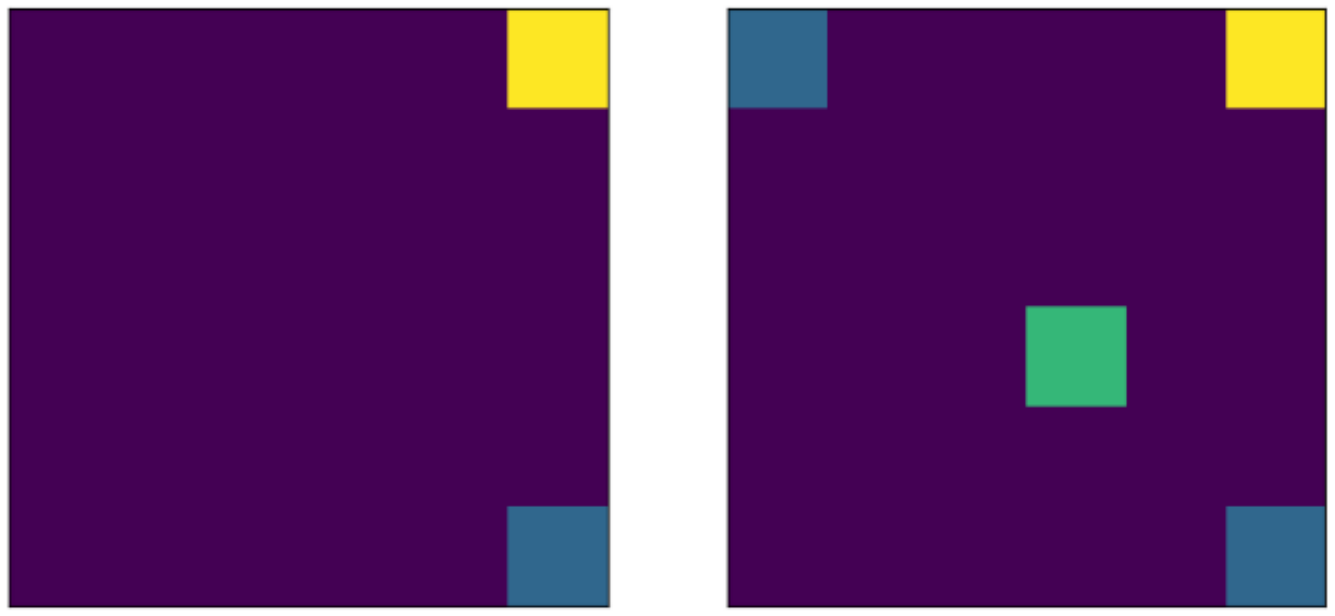

Fig. 13. Example of a 6x6px dummy image: Left: Ground-truth mask. Right: Result of a model prediction. The colors of the pixels refer to different classes.

In case of the yellow class, the intersection (TP) and union (TP + FP + FN) values are 1, since the pixel in the upper right corner is yellow in both images. In the case of the green class, the mask does not contain green pixels, so this class is discarded for the calculation. In the case of the blue class, the intersection value is 1 and the union is 2 (TP=1 and FP=1). For the background class, which consists of a union value of 34 pixels (TP=32 and FN=2), the intersection is 32, since there are only 2 pixels that were wrongly predicted. The IoU of each class is then calculated and averaged, discarding the green class.

As can be seen, the IoU of the green class is not considered, since there is no green object in the ground truth; however, the falsely predicted green pixel is considered as an incorrect value for the background class. The reason why we created this metric was because in some inferred masks we noticed some class pixels that were not in the ground-truth mask, due to poor EDF image quality or artifacts. In the following section, one can see the use of this metric and how it performs correctly.

## V. Results

From the evaluation of the 46 models prepared for the $Fe_{14}Nd_2B$-magnet experiment, we created different graphs to easily compare the results. As said before, the analysis should be focused on the use of the metric developed for this specific case, "Missing Class IoU", however, the accuracy and mIoU graphs are also presented to explain the results in more detail and to emphasize the use of the customized metric.

The scatter plots in this section were constructed with the same test dataset but with different metrics. The plots take as lower x-axis the percentage of synthetic data in addition to the original data, in other words, the results of multiplying the second and third column of TABLE III and TABLE IV, while the top x-axis represents the absolute amount of training data in the purely synthetic datasets (SyntLQ and SyntMQ), which are the same values as the amount of synthetic data added to the larger datasets (BLQ and BMQ).

### A. Analysis according to the percentage of synthetic data added

Fig. 14 shows the evaluation of the accuracy on the different datasets. It can be seen that the values of nearly all the models (except purely synthetic) were between 97.5% and 98%, with 100% being the maximum that could be reached. It can be observed that BQM [100%] had the highest value with a little more than 98% followed by the BMQ dataset with 50% of synthetic data and the SHQ [250%]. In the case of the models trained with purely synthetic data (brown bars with line pattern), the model with highest accuracy, 97.2%, is the one with 36 synthetic images. On the other hand, the models with the worst accuracy were in general those of SyntMQ and especially BMQ [300%] with a value of 91.4%, which is not shown in the graph to better observe the others. Furthermore, this graph allows one to visualize the metric in terms of the base models (horizontal colored and stylized lines), without synthetic data, where it can be seen that BLQ (green solid line) and SHQ (blue dotted line) had similar values around 98%.

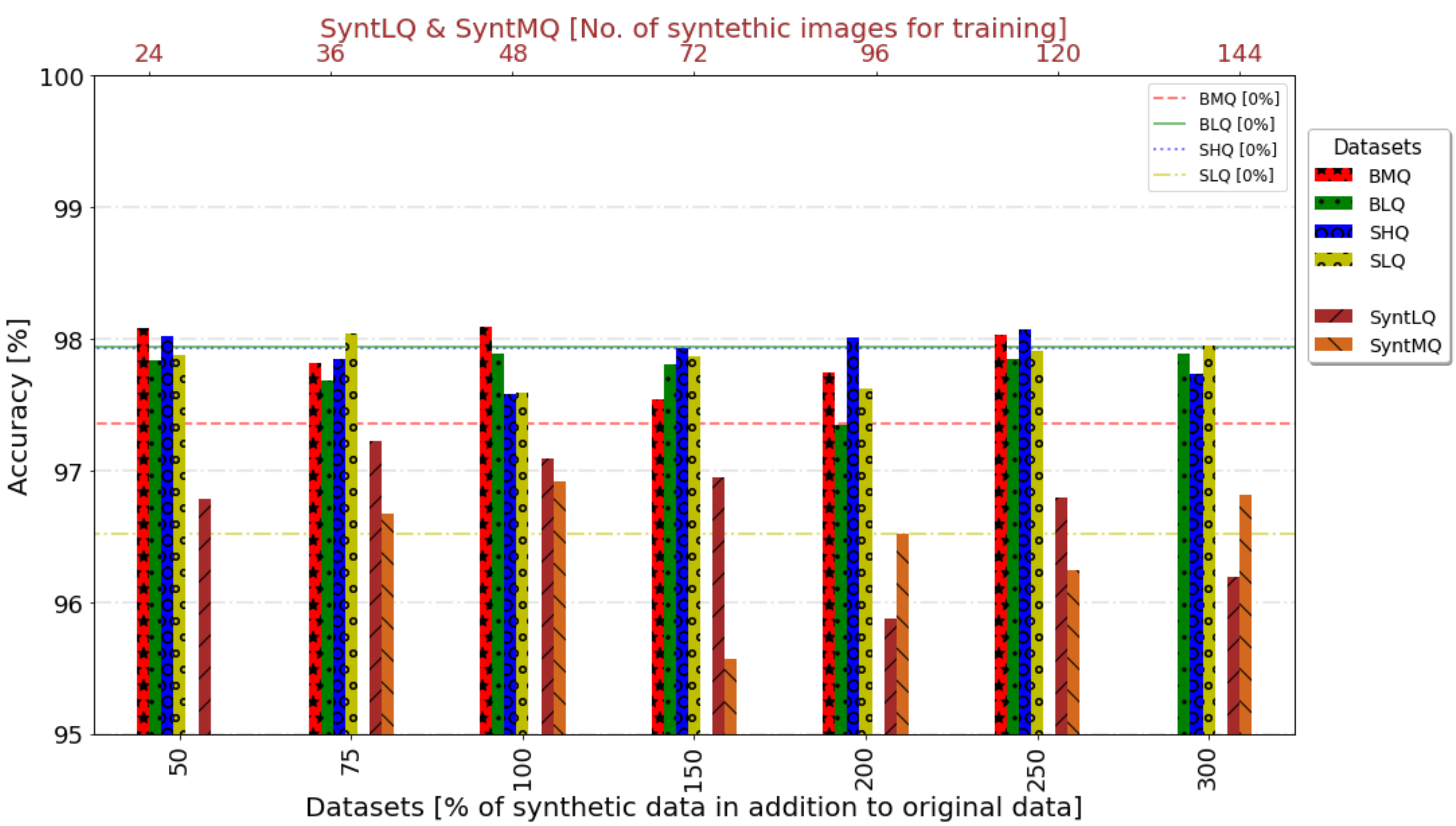

Fig. 14. Evaluation of accuracy on the 46 datasets according to the percentage of synthetic data added. The y-axis corresponds to the accuracy in %, the lower x-axis to the addition, in percentage, of synthetic data to the training datasets and the upper x-axis to the number of synthetic images, in the training dataset, for purely synthetic datasets. Stylized, colored horizontal lines represent the accuracy of the base models (without addition of synthetic data) and each bar represents the accuracy of the models trained with addition of synthetic data. Note that BMQ [300%] and SyntMQ [50%] are not shown in the graph because their accuracy was less than 95%.

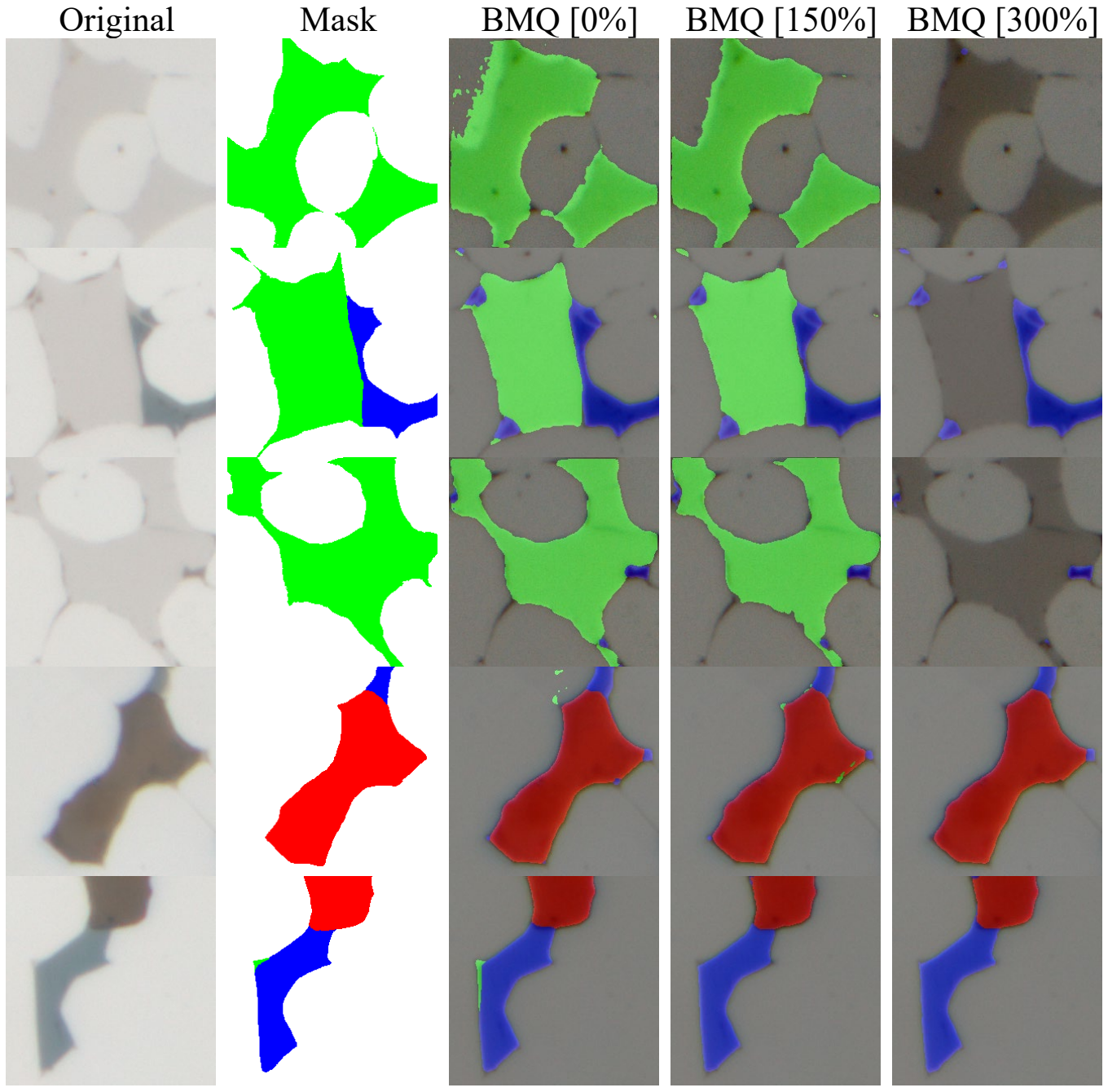

Fig. 16. $Fe_{14}Nd_2B$-magnet - Segmentation on five images from the test dataset. By column: 1. Original microstructural image; 2. Corresponding mask in RGB; 3. Segmented image with the BMQ base model overlapped to the original; 4. Segmented image with the BMQ model [150%] overlapped to the original; 5. Segmented image with the BMQ model [300%] overlapped to the original.

In the mIoU graph (Fig. 16), the values are distributed over a wider range than the accuracy graph, for example, for the BMQ (red bars with stars pattern), its range goes from 69% mIoU with the base model (red dashed lines) to 82% with the model trained with 50% synthetic data. Furthermore, if BMQ [0%] is excluded from the graph, the BMQ model with the lowest mIoU would be the one with 150% synthesized data (mIoU=70.5%), while the one with the most synthesized data (300%) has an mIoU of 71.6%.

The plot also shows that, in some cases, the mIoU values of the models with pure synthetic data were higher than those of the models trained with combined data. Additionally, when adding 50% synthesized data to the base models, both SLQ and SHQ had a decrease in their metrics, while the opposite happened to BLQ and BMQ.

After highlighting the main results generated by the above metrics, we carried out the analysis of the "Missing Class IoU" for each dataset and, in turn, the graphs of this metric for each class separately.

The main observations from Fig. 17 are as follows:

- By adding 50% of synthetic data, except for the pure synthetic data, all the models improved.
- The purely synthetic SyntLQ model with 24 images had a higher value (87.3%) of the metric than the base SLQ model (86.1%).

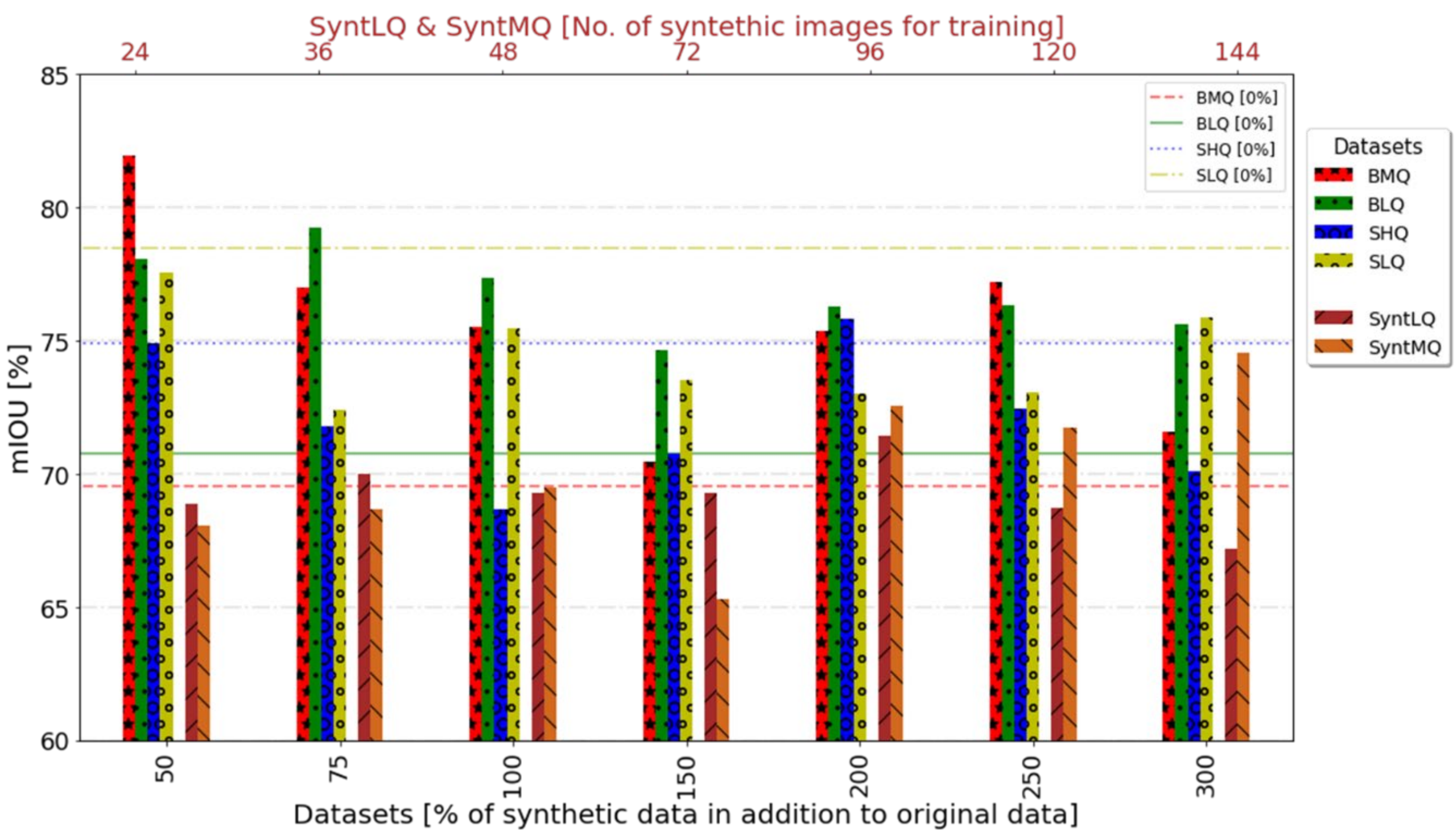

Fig. 15. Evaluation of mIoU on the 46 datasets according to the percentage of synthetic data added.

- BMQ [100%] had the best performance (91%).
- BMQ [300%] had the worst performance (76.1%), barely lower than SyntMQ [50%].
- SLQ [250%] almost reached the best performance of SHQ (90.3%) and the performance of both at [50%] was similar (≈90%).
- All models (except the purely synthetic ones) achieved an improvement over the initial state in almost all datasets extended with synthetic data.
- Models without synthetic data have a value of about 89%, except for the SLQ with 86.1%.
- Models trained with purely synthetic data are located below all those combined by real and synthetic data.

Analysis per class/phase

We extracted the values from the "Missing Class IoU" for each phase to provide another point of view. For example, when analyzing the best model, BMQ [100%], the metric improved for absolutely all classes compared to the base model, as can also be seen in the inferred images (Fig. 15). Moreover, when all of the graphs are examined together, for all phases except Nd-rich, the models trained with purely synthetic data are below the others, for each different value of the x-axis.

In the case of the first graph, that of the matrix phase, one can see that the values are mostly compressed between 97% and 98% approximately, followed by the graph of the oxide phase evaluation, between 90.5% and 92.5%. Then come those of Nd-rich with values lower than 85% and Eta, below 80%. Some low values are also observed (excluding models trained with purely synthetic data), e.g., in the case of SLQ [0%] compared to their respective models trained on synthetic data, and in the case of BMQ [300%], on the evaluation of the matrix and Eta phases.

## VI. Discussion

Based on the results, it is possible to improve semantic segmentation models by using synthetic data. This approach allows to decrease the time consumption in labeling large datasets as well as populating small ones.

It could be seen why accuracy and mIoU should not be used in this specific experiment with $Fe_{14}Nd_2B$-magnet images. The first metric, mainly because almost all models had a value higher than 97.5%; this would reflect that they were almost perfect, whereas this was not the case, as could be seen with the other metrics and the predicted images (Fig. 15).

For the second metric, a clear case of its limitation in this analysis is, for example, the following. When comparing the mIoU values of the BMQ [150%] and BMQ [300%] models, a first intuition would be that the latter performed better than the former; however, when observing again those predicted images, it can be seen that the Eta phase was not detected for the latter model.

To validate the development of the "Missing Class IoU",

Fig. 19 shows two clear examples with predicted masks using the BMQ model with 150% and 300% synthetic data. For the images in the first row, it can be seen that the results of the metrics are the same for both models, since all the values of each class are taken into account.

However, in the second example, the difference is that the green class is not considered for the calculation of the image inferred by BMQ [150%], so the sum of the IoU per class is divided by 2 and not by 3 (3 because the red class is not present). A final clarification on this situation is that, if someone does not see the inferred images and receives the information that the mIoU of an image is 65%, one might think

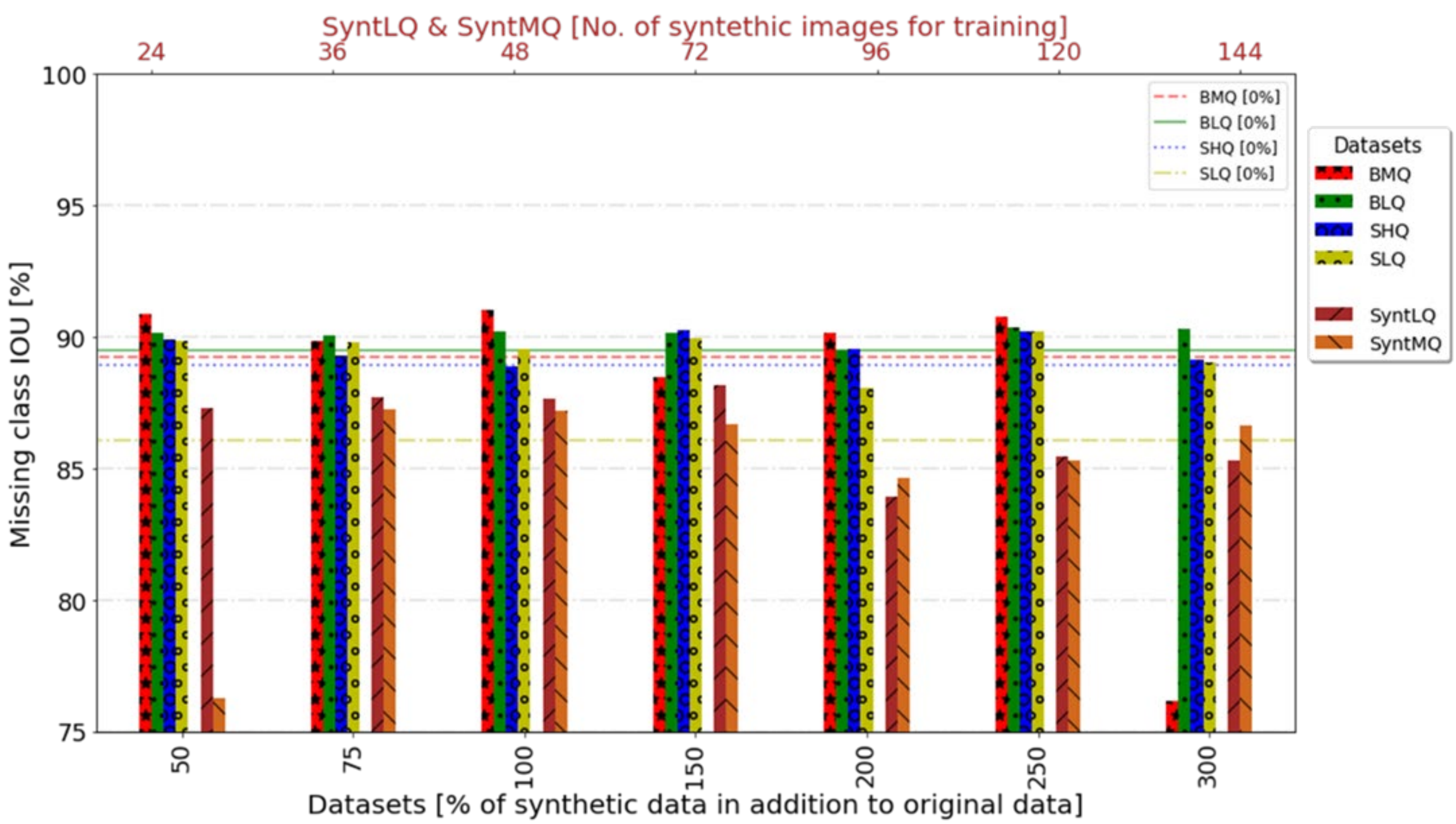

Fig. 17. Evaluation of "Missing Class IoU" on the 46 datasets according to the percentage of synthetic data added.

that the model has completely failed in its prediction, when, in fact, it has not.

In addition, if the BLQ and BMQ base models are compared with their respective ones by adding 50% of synthetic data, the improvement is very large but this can be confirmed by looking at the inferred images by these models or plotting the mIoU of each image on a histogram, as something similar to the above may occur.

Although it is not clear how much data to generate and use, the combination of both original and synthetic data produces better models than both separately. Besides, this can be checked by generating a large amount of synthetic data, then training a few models by taking different percentages and observing the "Missing Class IoU" graph.

Thus, in the current analysis, the BMQ [100%] model would be chosen, which has increased from 89.5% to 91% due to the addition of synthetic data. Fig. 20 shows some inferred images by this model. Note that in case of other materials and approaches, the appropriate metric should be chosen for the selection of the best model.

Moreover, the use of "per class/phase" graphs allows the analyst to observe poor inference in specific classes, commonly due to class imbalance or labeling errors in the datasets. In this way, synthetic images with more composition of the unbalanced class and more accurate masks can be added simultaneously.

Other conclusions can also be drawn from these graphs, e.g. the first two plots from Fig. 18 (matrix, Eta) show that the evaluation is consistent, for instance, by checking the SLQ [0%] and BMQ [300%] which had the lowest values of their corresponding datasets. Both models had problems in differentiating the Eta phase from the Matrix.

On the other hand, the large improvement of SLQ with 50% synthetic data was basically due to improved detection of the Eta phase, while the worsening of BMQ from 250% to 300% was due to the null detection of that phase.

Regarding the quality of the images, one aspect to note in the production of the 4-channel image is that, although the VQ-VAE and PixelCNN models are trained with the BMQ, which has several labeling errors, the synthetically produced masks have a high accuracy and show almost no pixel value errors. Whereas in reality, in most cases, one does not have perfect masks for training (Fig. 21).

The important question to answer would be why the semantic segmentation models were improved with synthetic images, to which a possible answer would be the fact that the generated labels are more accurate. Although the generated microstructure does not look 100% real, as described in the literature, labeling errors lead to a worse performance of the models.

When one creates a semantic segmentation model and its results are not as desired, the first thing to look at is the conditions of the masks. If these are correct, instead of acquiring more images and labeling them, it may be a great option to use the workflow explained in this thesis, since on the same day one can train the set of models and generate synthetic

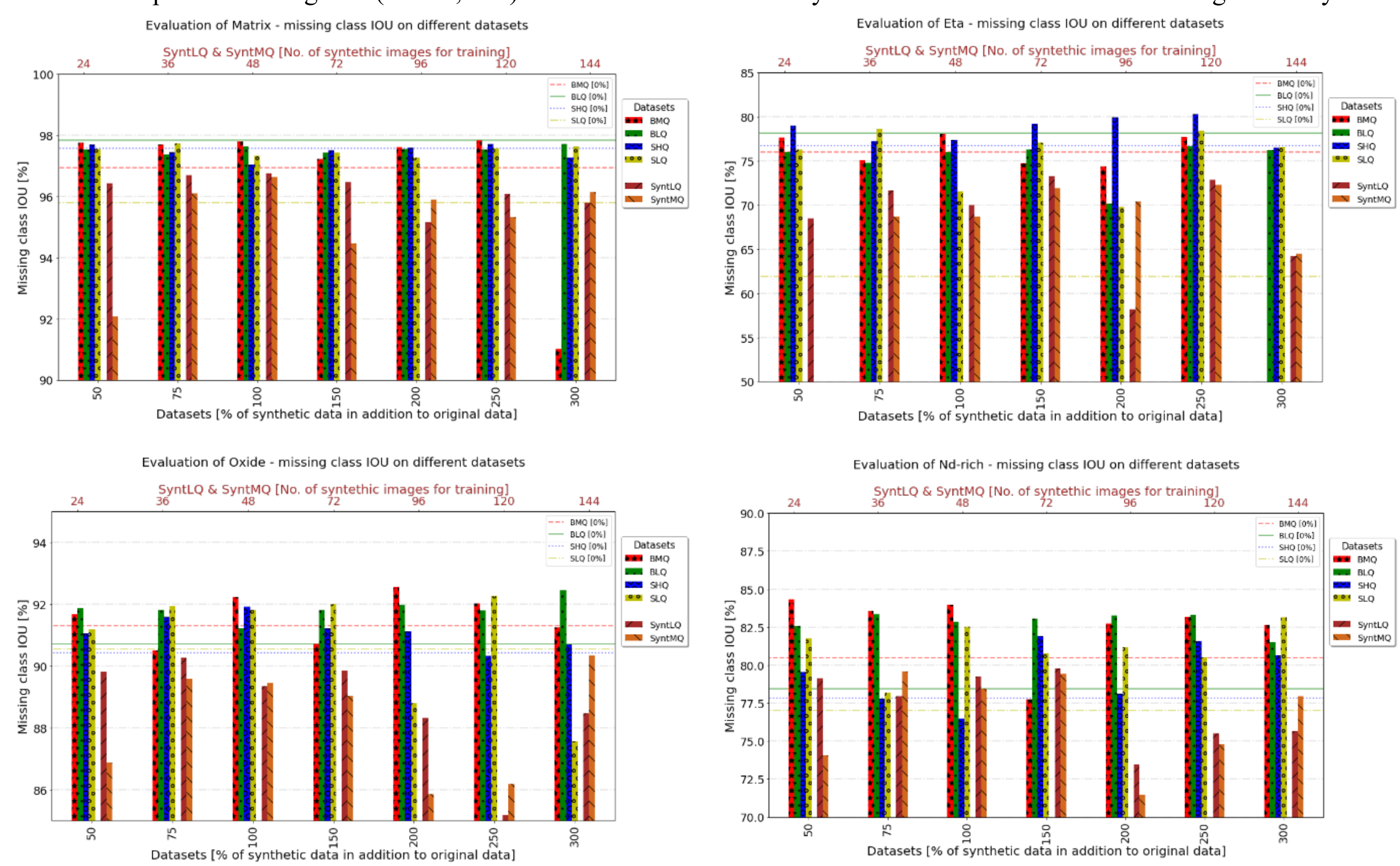

Fig. 18. Evaluation of "Missing Class IoU" per class/phase on the 46 datasets. Note that BMQ [300%] and SyntMQ [50%] are not shown in the "Evaluation of Eta" graph because their corresponding value for the metric were 0% and 14.9%.

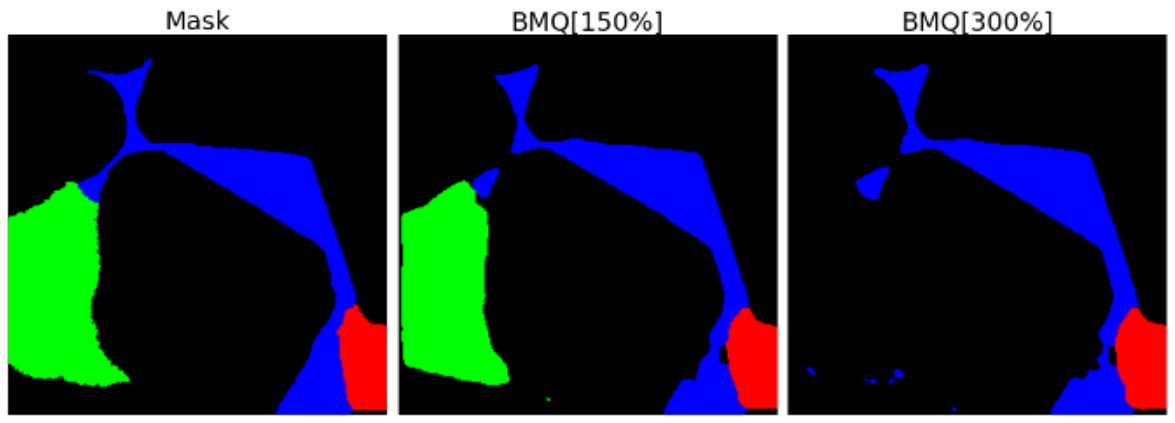

| | [%] | [%] |
|---|---|---|
| **IoU**$_{Black}$ | 96.7 | 84.7 |
| **IoU**$_{Blue}$ | 86.5 | 83.0 |
| **IoU**$_{Red}$ | 93.2 | 92.5 |
| **IoU**$_{Green}$ | 90.9 | 0.0 |
| **mIoU** | 91.8 | 65.1 |
| **Missing Class IoU** | 91.8 | 65.1 |

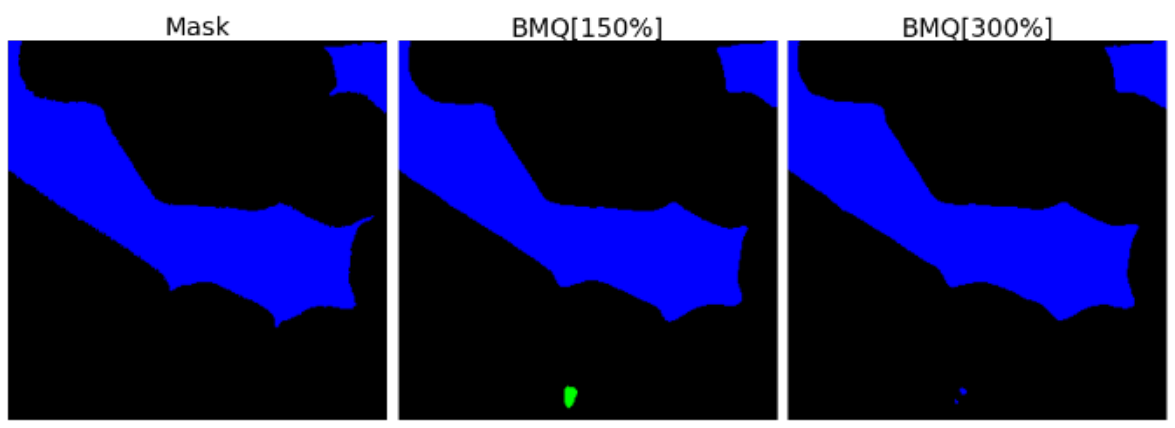

| | [%] | [%] |
|---|---|---|
| **IoU**$_{Black}$ | 98.6 | 99.0 |
| **IoU**$_{Blue}$ | 96.4 | 96.9 |
| **IoU**$_{Red}$ | Nan | Nan |
| **IoU**$_{Green}$ | 0.0 | Nan |
| **mIoU** | 65.0 | 98.0 |
| **Missing Class IoU** | 97.5 | 98.0 |

Fig. 19. Two examples to understand the "Missing Class IoU" metric. By column: 1. Original RGB mask. 2. Mask predicted with the BMQ [150%] model. 3. Mask predicted with the BMQ [300%] model

images that could improve the segmentation, all much faster than the time needed in sample preparation, image acquisition and labeling, thus saving time and costly tasks.

One more point should be noted, which is that of data augmentation. Applying, for example, the Albumentations library [19] to augment the datasets requires some work. One must correctly define which image transformation operations to be used. For example, a U-Net model was trained with the BMQ dataset without synthetic data and with some data augmentation, e.g., contrast, brightness, blur, etc., which was not adequately verified for this type of dataset. As a result, a worse performance (Missing Class IoU = 89.2%) than the base BMQ (Missing Class IoU = 89.5%) was obtained.

Although with the correct application of data augmentation, a great performance could probably be achieved and maybe even better than BMQ [100%], the application of the workflow of this paper does not involve a previous study of the images. In this case, there is no need to define what kind of augmentation to apply but only requires combining the real image with the mask and training the VQ-VAE model.

Depending on the distribution of the codes, it is the model that will be used to generate new codes; for example, during experimentation, PixelCNN has shown good results for non-uniform microstructures.

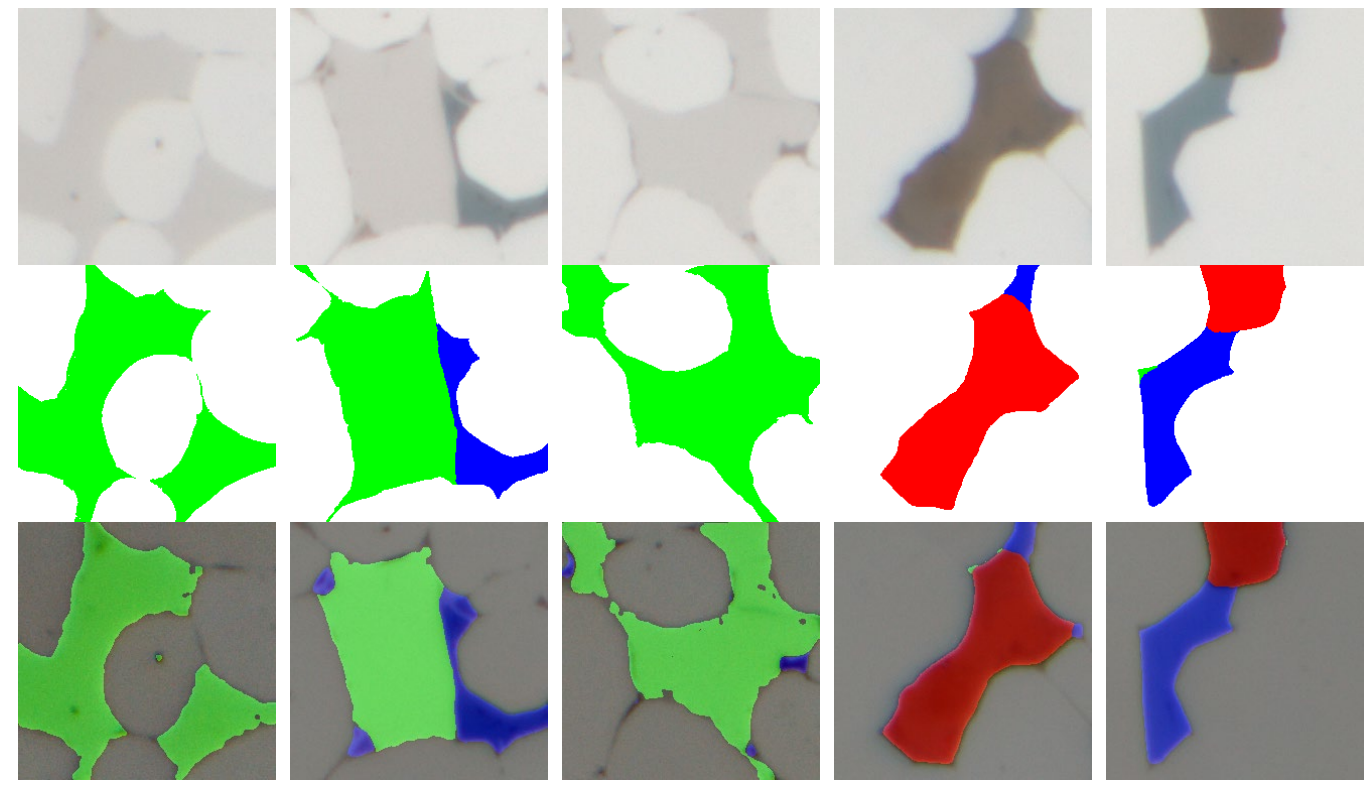

Fig. 20. Segmentation on five images from the test dataset. By row: 1. Original microstructural image; 2. Corresponding mask in RGB; 3. Segmented image with the BMQ [100%] model and overlapped on the original.

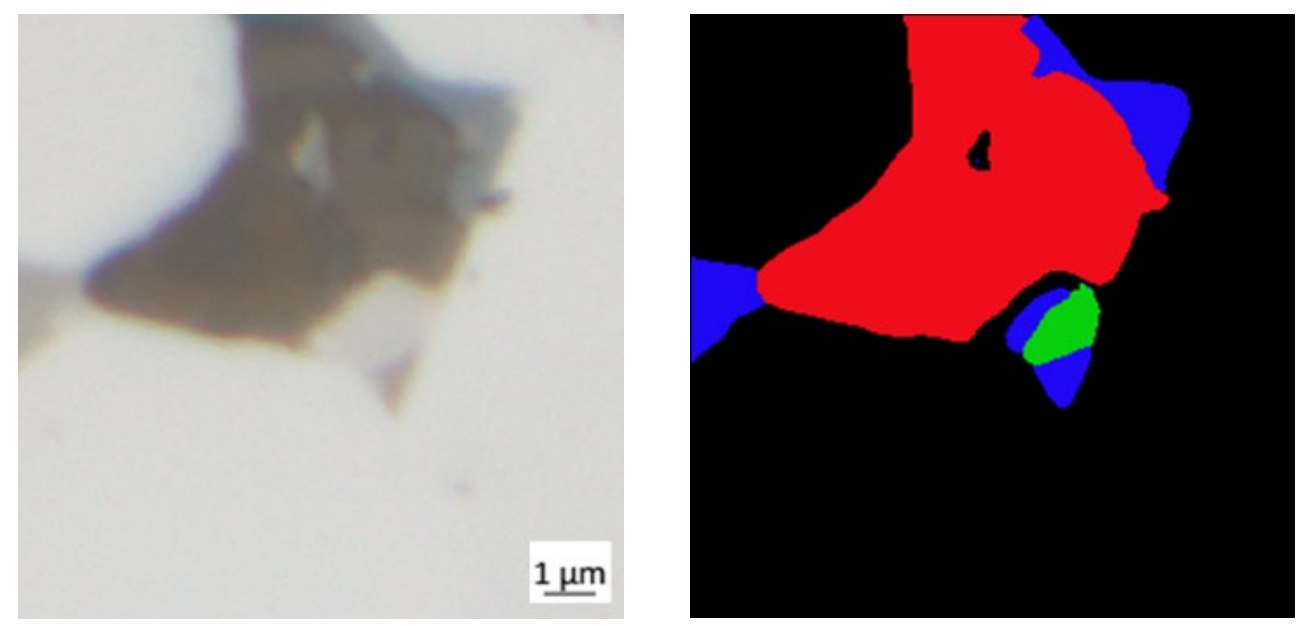

Fig. 21. $Fe_{14}Nd_2B$-magnet image patch taken from the training dataset. The microstructural image has unsharp boundaries, variation of pixel colors due to EDF (blue pixels can be seen in the oxide phase) and a huge wrong label.

## VII. Limitations of the Current Workflow

Although the images generated for the experiment presented in this paper served to improve the semantic segmentation models, they were not perfectly morphologically accurate and were somewhat blurred.

The former was due to the PixelCNN model, since the z-matrices obtained by the VQ-VAE model differentiated each phase well. Further experiments were performed with other materials, e.g., Li-ion battery microstructures, in which the microstructure layers are more clearly distributed, and the result of the PixelCNN model was poor, as it failed to learn the distribution of the layers.

The second problem is the reconstruction of blurred images due to the VQ-VAE model, which is explicitly shown in the authors' paper.

## VIII. Conclusion

In conclusion, we would like to highlight a few points:

- Using the combination of both original and synthetic data results in improved performance (1-4%) of the semantic segmentation model compared to training with both separately.
- The synthetically produced masks have well-defined object edges and barely show any over segmentation or misclassification of pixels, even if the models were trained with various labeling errors.
- VQ-VAE exhibited good performance but should still produce less blurred images. PixelCNN is adaptable to approaches, where the phases are more sparsely distributed and not very adaptable in case of densely concentrated regions, such as in a pattern form where the pixels are distributed in a structured way. Nevertheless, there are new versions of PixelCNN, that improve generation performance and overcome the "blind spot problem", that can be tested for this workflow.

- The proposed "Missing Class IoU" is useful for evaluation in cases where images have small objects that were not labeled but detected, and whose class is not part of the label. Thus, very low mIoU values are avoided.
- "Per class/phase" graphs allow the analyst to observe poor inference in specific classes (e.g. 14.9% on the evaluation of "Missing Class IoU" on Eta phase with SyntMQ [50%]) and a more detailed understanding of the metrics used.
- For a significant improvement of the workflow, it would be of interest to generate the dual synthetic image with the incorporation of variables in the latent space to be able to manipulate microstructural properties, e.g. phase fraction, in order to populate or balance datasets with the desired phase/property.
- This approach would overcome a limitation of data augmentation, which focuses on creating new images by making global changes to the original image. The augmented data do not have semantic meaning and therefore, do not improve generalization ability of the semantic segmentation models when there is imbalance in different classes. Carrying out local changes or adding microstructural information in the latent space could overcome imbalance of specific classes.

## Acknowledgment

This work was funded by the Federal Ministry of Education and Research Germany (BMBF) in the scope of the WirLebenSOFC project (grant no. 03SF0622C). The authors would also like to thank Amit Kumar Choudhary and Patrick Krawczyk for their valuable discussions and feedback.